\pdfoutput=1

\documentclass[11pt]{article}

\usepackage[]{ACL2023}
\usepackage{times}
\usepackage{latexsym}
\usepackage{supertabular}
\usepackage{graphicx}
\usepackage{stfloats}
\usepackage{tabularx}
\usepackage{multirow}
\usepackage{makecell}
\usepackage{booktabs}
\usepackage{amsmath}
\usepackage{amsfonts}
\usepackage{amssymb}
\usepackage{enumerate}
\usepackage{algorithm}
\usepackage{algorithmic}
\usepackage{bbm}
\usepackage{xcolor}
\usepackage{color}
\usepackage{pifont}

\usepackage{times}
\usepackage{latexsym}

\usepackage[T1]{fontenc}

\usepackage[utf8]{inputenc}

\usepackage{microtype}

\usepackage{inconsolata}

\def\benchmark{TransportationGames} 

\title{\benchmark: Benchmarking Transportation Knowledge of (Multimodal) Large Language Models}

\author{Xue Zhang, Xiangyu Shi, Xinyue Lou, Rui Qi,\\
\textbf{Yufeng Chen, Jinan Xu, Wenjuan Han\thanks{\enspace Wenjuan Han is the corresponding author.}}\\
        Beijing Jiaotong University, Beijing, China \\ 
        \texttt{\{23111135,22120416,20241254,20281284,chenyf,jaxu,wjhan\}@bjtu.edu.cn}}

\begin{document}
\maketitle
\begin{abstract}
Large language models (LLMs) and multimodal large language models (MLLMs) have shown excellent general capabilities, even exhibiting adaptability in many professional domains such as law, economics, transportation, and medicine.
Currently, many domain-specific benchmarks have been proposed to verify the performance of (M)LLMs in specific fields.
Among various domains, transportation plays a crucial role in modern society as it impacts the economy, the environment, and the quality of life for billions of people.
However, it is unclear how much traffic knowledge (M)LLMs possess and whether they can reliably perform transportation-related tasks.
To address this gap, we propose {\benchmark}, a carefully designed and thorough evaluation benchmark for assessing (M)LLMs in the transportation domain.
By comprehensively considering the applications in real-world scenarios and referring to the first three levels in Bloom's Taxonomy, we test the performance of various (M)LLMs in memorizing, understanding, and applying transportation knowledge by the selected tasks.
The experimental results show that although some models perform well in some tasks, there is still much room for improvement overall.
We hope the release of {\benchmark}\footnote{The evaluation code will be released soon in \url{http:/transportation.games}.} can serve as a foundation for future research, thereby accelerating the implementation and application of (M)LLMs in the transportation domain.

\end{abstract}

\section{Introduction}
Large language models (LLMs) are revolutionizing the way humans work by augmenting them in various tasks. As these LLMs, for example GPT-4~\cite{openai2023gpt4} and LLaMA~\cite{touvron2023llama}, become more sophisticated, they will be able to handle more complex tasks, enabling them to assist and collaborate with humans in a multitude of professional domains~\citep{sanh2021multitask,ouyang2022training, zhang2022opt, shao2023compositional}. 
Additionally, beyond single LLM, the Multimodal Large Language Model (MLLM) has recently emerged as a popular area of research. The MLLM utilizes powerful LLMs to effectively handle multimodal tasks, resulting in versatile problem solvers.
To comprehensively and accurately assess the capabilities of (M)LLMs, evaluation benchmarks play a crucial and indispensable role in their development~\cite{hendrycks2020measuring}. By evaluating (M)LLMs using these benchmarks, researchers and developers can gain valuable insights into the strengths and weaknesses of different models, enabling them to identify areas for improvement and innovation. 

Currently, many benchmarks have been proposed to assess (M)LLMs on various aspects of universal capabilities, \textit{e.g.}, MMLU~\cite{hendrycks2020measuring}, C-Eval~\cite{huang2023ceval}, CMMLU~\cite{li2023cmmlu}, BIG-bench~\cite{srivastava2023beyond}, MMBench~\cite{liu2023mmbench} and MME~\cite{fu2023mme}.
Moreover, when evaluating (M)LLMs, it is important to not only focus on their general capabilities but also to incorporate domain-specific benchmarks for assessing models specialized in specific fields~\cite{zhao2023domain}, because domain-specific benchmarks push (M)LLMs towards tackling the specific challenges and complexities of their target fields, ultimately driving practical progress and responsible implementation. 
Existing domain-specific benchmarks include LawBench~\cite{fei2023lawbench}, LegalBench~\cite{guha2023legalbench}, and LAiW~\cite{dai2023laiw} for the legal domain, MIR-based benchmark~\cite{goenaga2023explanatory} for the medicine domain, ChemLLMBench~\cite{guo2023large} for the chemistry domain, etc.
Among various domains, transportation plays a crucial role in modern society as it impacts the economy, the environment, and the quality of life for billions of people~\cite{taylor2015transportation,koopmans1949optimum}. 
However, it is unclear how much traffic knowledge\footnote{We only focus on Chinese.} (M)LLMs possess and whether they can reliably perform transportation-related tasks.

\begin{figure*}[t]
    \centering
    \includegraphics[width=\textwidth]{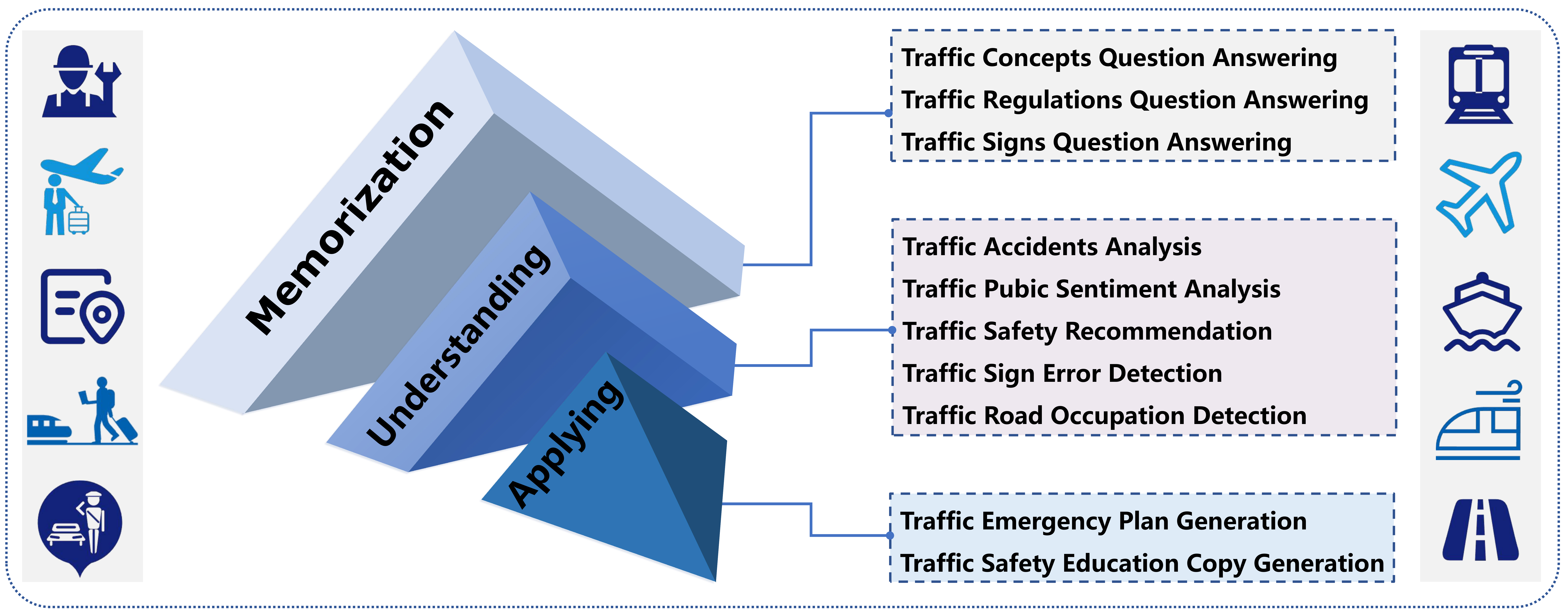}
    \caption{
    The organization of our \benchmark.
    Considering the specific scenarios in the transportation domain, our {\benchmark} employs the first three levels in Bloom's Taxonomy, which are Memorization, Understanding, and Applying, to evaluate the (M)LLMs. We select 10 tasks based on diverse sub-domains in the transportation domain such as urban transportation, rail transit, aviation, and maritime transport.
    }
    \label{fig:benchmark}
\end{figure*}

To address this gap, we introduce {\benchmark} (refer to Figure \ref{fig:benchmark}): a thoughtfully designed, all-encompassing evaluation benchmark to accurately evaluate the capabilities of (M)LLMs in executing transportation-related tasks. 
By comprehensively considering the applications in real-world scenarios, we select 10 varied tasks across 3 types: multiple-choice, ``True/False'' judge, and text generation. 
We categorize these tasks into three skill levels based on widely recognized Bloom's cognitive models~\citep{krathwohl2002revision}: (1) Transportation knowledge memorization: whether (M)LLMs can memorize transportation-relevant concepts, facts, regulations, and traffic law articles; (2) Transportation knowledge understanding: whether (M)LLMs can understand, analyze and reasoning in transportation articles; (3) Transportation knowledge applying: whether (M)LLMs can effectively make the necessary logical deductions to solve practical transportation tasks both for public and professionals. 
Overall, our {\benchmark} offers a systematic outline of the skillset necessary for tasks related to transportation.

Our main contributions are three-fold:
\begin{itemize}
    \item \textbf{Systematically-constructed benchmark.} We introduce \benchmark, a carefully designed and thorough evaluation benchmark for assessing (M)LLMs in transportation-related tasks. It is the first benchmark specifically designed for the transportation domain. 
    \item \textbf{Experiments.} 
    We design appropriate rules to accurately extract answers from the model-generated predictions, and employ proper metrics for each task. We conduct extensive testing on 16 widely used (M)LLMs and the evaluation results are presented in Table \ref{table:LLMs-res} and Table \ref{table:MLLMs-res}.

   \item \textbf{Analysis.} 
   We observe that although some LLMs perform well in some tasks on text-only knowledge, there is still room for improvement. As for multimodal knowledge, most MLLMs exhibit poor capability.
   Additionally, we analyze the key factors affecting model performance. 
\end{itemize}

\section{Related Work}
\subsection{Large Language Models}
Large language models (LLMs) typically refer to Transformer language models encompassing several billion (or more) parameters~\cite{LLMSurvey}, such as GPT-4~\cite{openai2023gpt4},  LLaMA~\cite{touvron2023llama}, Baichuan~\cite{yang2023baichuan}, and so on. These models undergo training on extensive corpora of textual data, thereby acquiring excellent capabilities.
The comprehensive training processes for LLMs encompass stages such as model pretraining, instruction tuning, reward model training, and reinforcement learning with human feedback (RLHF). Through the implementation of these training strategies, LLMs can achieve commendable performance on tasks within general domains. 
To improve the performance of LLMs on more specific domains, more research endeavors increasingly aspire to deploy LLMs across diverse domains, including but not limited to law, medicine, transportation, chemistry, and psychology, to proficiently accomplish domain-specific tasks.
Moreover, multimodal large language models (MLLMs) have emerged as a recent focal point in the community, capitalizing on the prowess of potent large language models to serve as cognitive entities for executing multimodal tasks, thereby exhibiting remarkable emergent capabilities.

In this paper, we focus on the development of (M)LLMs in the transportation domain. 
There are many (M)LLMs tailored for the traffic domain including TransGPT(-MM)~\cite{TransGPT, TransGPTMM}, TrafficGPT\footnote{\url{https://github.com/lijlansg/TrafficGPT}}, MT-GPT\footnote{\url{https://www.7its.com/?m=home&c=View&a=index&aid=19245}}, and TransCore-M\footnote{\url{https://github.com/PCIResearch/TransCore-M}}.
Among them, TransGPT(-MM) and TransCore-M have undergone instruction tuning based on traffic domain data.

\subsection{Existing Benchmarks}
The comprehensive and precise evaluation of the functionalities inherent in (M)LLMs is pivotal and irreplaceable in their development. Evaluation benchmarks assume a critical role in this context, furnishing a standardized framework that facilitates the meticulous measurement and analysis of (M)LLM performance across diverse tasks and domains. The scrutiny of (M)LLMs through these benchmarks endows researchers and developers with invaluable insights into the nuanced strengths and vulnerabilities characterizing distinct models. In turn, this empowers them to discern specific domains necessitating refinement and innovation, thereby enhancing the overall development of (M)LLMs.

Recently, more and more benchmarks have been developed to evaluate the various capabilities of (M)LLMs.
To assess the comprehensive capabilities of LLMs, many benchmarks have been constructed based on knowledge spanning various disciplines and languages, including MMLU~\cite{hendrycks2020measuring} and ARC~\cite{clark2018think}, which are grounded in English, as well as C-Eval~\cite{huang2023ceval} and CMMLU~\cite{li2023cmmlu}, which are rooted in Chinese.
As for MLLMs, there are also many benchmarks with the comprehensive evaluation pipeline, such as MME~\cite{fu2023mme} and MMBench~\cite{liu2023mmbench}. 
In addition, some benchmarks are designed to evaluate the performance of (M)LLMs on some specific domains, \textit{e.g.}, LawBench~\cite{fei2023lawbench}, LegalBench~\cite{guha2023legalbench}, and LAiW~\cite{dai2023laiw} for the legal domain, MIR-based benchmark~\cite{goenaga2023explanatory} for the medicine domain, ChemLLMBench~\cite{guo2023large} for the chemistry domain, and so on.
However, to the best of our knowledge, there is no systematic evaluation benchmark for the transportation domain, so we propose the {\benchmark} for assessing (M)LLMs in transportation-related tasks.

\begin{table*}[t]
\centering
\resizebox*{\linewidth}{!}{
\begin{tabular}{l|clccc}
\bottomrule
\textbf{Capability Levels} & \textbf{ID} & \textbf{Task} & \textbf{Modality} & \textbf{Type} & \textbf{Metric} \\
\hline
\multirow{3}{*}{\textbf{\makecell[l]{Transportation Knowledge\\Memorization}}} & \textbf{T1} & Traffic Concepts Question Answering & Text & TF/MLC & Accuracy \\
& \textbf{T2} & Traffic Regulations Question Answering & Text & TF/MLC & Accuracy \\
& \textbf{T3} & Traffic Signs Question Answering & Multimodal & TF/MLC & Accuracy \\

\hline
\multirow{5}{*}{\textbf{\makecell[l]{Transportation Knowledge\\Understanding}}} & \textbf{T4} & Traffic Accidents Analysis & Text/Multimodal & Generation & ROUGE/GPT-4 \\
& \textbf{T5} & Traffic Pubic Sentiment Analysis & Text & Generation & ROUGE/GPT-4 \\
& \textbf{T6} & Traffic Safety Recommendation & Text/Multimodal & Generation & ROUGE/GPT-4 \\
& \textbf{T7} & Traffic Sign Error Detection & Multimodal & Generation & ROUGE/GPT-4 \\
& \textbf{T8} & Traffic Road Occupation Detection & Multimodal & Generation & ROUGE/GPT-4 \\

\hline
\multirow{2}{*}{\textbf{\makecell[l]{Transportation Knowledge\\Applying}}} & \textbf{T9} & Traffic Emergency Plan Generation & Text/Multimodal & Generation & ROUGE/GPT-4 \\
& \textbf{T10} & Traffic Safety Education Copy Generation & Text & Generation & ROUGE/GPT-4 \\

\toprule
\end{tabular}
}
\caption{\label{table:tasks}
Task list of \benchmark. There are 10 tasks corresponding to 3 capability levels: Transportation Knowledge Memorization, Understanding, and Applying, and 2 modalities: Text and Multimodal (text + image), and 3 task types: multiple-choice (MLC), ``True/False'' judge (TF), and text generation. Additionally, the metrics used for each task are also listed and described in detail in \S\ref{sec:3-evaluation}.
}
\end{table*}

\section{Benchmark Construction}
In this section, we provide a detailed introduction to the construction of our \benchmark. 
Firstly, we elucidate the classification criteria (\S\ref{sec:3-taxonomy}) employed in the design of the benchmark, along with the corresponding selection of evaluation tasks (\S\ref{sec:3-tasks}). 
Subsequently, we introduce the data collection procedures (\S\ref{sec:3-data}) and the adoption of evaluation metrics (\S\ref{sec:3-evaluation}).

\subsection{The Taxonomy of \benchmark}
\label{sec:3-taxonomy}
In the construction of benchmarks, an effective process involves not only evaluating models on multiple sub-tasks but also organizing benchmarks systematically. These benchmarks can be organized based on task difficulty or task categories, providing a nuanced reflection of the models' aptitude. 
However, such a simplistic classification criterion may not adequately capture the full range of model capabilities.

Inspired by ~\citet{fei2023lawbench}, we adopt Bloom's cognitive model for task classification, aiming to capture the models' capabilities at a higher level.
Bloom's Taxonomy system, initially introduced by the educational psychologist Benjamin Bloom and his collaborators in 1956, has obtained widespread application and continuous development in subsequent decades. It has proven instrumental in assisting educators in both curriculum design and the evaluation of student learning outcomes. The taxonomy categorizes learning objectives within the cognitive domain into six progressively ascending levels: Remember, Understand, Apply, Analyze, Evaluate, and Create. These hierarchical levels delineate the depth and intricacy of cognitive learning, providing educators with a structured framework for instructional design and assessment.

Considering the specific scenarios in the transportation domain, we employ the first three levels in Bloom's Taxonomy to assess the (M)LLMs as shown in Figure \ref{fig:benchmark}. The detailed introduction is as follows:

\paragraph{Transportation Knowledge Memorization.} It tests whether (M)LLMs can memorize and answer basic transportation-related knowledge, such as concepts, facts, regulations, or traffic law articles.

\paragraph{Transportation Knowledge Understanding.} 
The excellent understanding capability generally requires the model to engage in activities such as interpretation, illustration, categorization, summarization, and inference based on transportation-domain knowledge.
For example, the models can interpret traffic regulations and rules, compare the applicable conditions of different rules, classify traffic rules based on some features, etc.

\paragraph{Transportation Knowledge Applying.}
The applying capability is to assess whether the model can flexibly apply acquired knowledge to solve practical transportation tasks both for the public and professionals.

\subsection{Tasks}\label{sec:3-tasks}
The core knowledge areas of the transportation profession generally include transportation infrastructure construction, carrier theory and technical equipment, transportation system planning, port and station hub planning and design, passenger operation organization, cargo operation organization, operation dispatching command, as well as transportation policies and regulations, transportation commerce, transportation economy, transportation safety, modern logistics, and comprehensive transportation.
And it mainly involves four sub-domains: road transportation, railway transportation, waterway transportation, and aviation transportation.

During selecting tasks, we take into account diverse sub-domains in transportation and the varying needs of different people, including the general public and industry practitioners, in their day-to-day lives or professional undertakings. Furthermore, we conduct detailed consultations with domain experts to choose the specific tasks.
Finally, we select 10 tasks under the aforementioned capability levels and the task list is presented in Table \ref{table:tasks}.
Note that due to the different application scenarios of different tasks, it involves multiple modalities of knowledge, such as text and image modality.
The concrete introduction is as follows.

\paragraph{Transportation Knowledge Memorization Tasks}
\begin{itemize}
    \item \textbf{Traffic Concepts Question Answering (T1)}: Inquire about common concepts in the field of transportation, formulating queries in either multiple-choice (MLC) or ``True/False'' judge (TF) formats. In the case of multiple-choice questions, the model is expected to select the correct answer from a set of four options, whereas true/false questions necessitate the model to determine the correctness of a given statement.
    \item \textbf{Traffic Regulations Question Answering (T2)}: Question the model regarding nuanced components of traffic regulations, including numerical parameters, years, or analogous elements. The question formats are MLC or TF.
    \item \textbf{Traffic Signs Question Answering (T3)}: Given a traffic sign image and a query, test whether the model can memorize the meaning of different traffic signs. The query formats are MLC or TF.
    
\end{itemize}

\paragraph{Transportation Knowledge Understanding Tasks}
\begin{itemize}
    \item \textbf{Traffic Accidents Analysis (T4)}: Given a photo of a traffic accident scene or a traffic accident course, the model is required to extract and summarize information including the origins, progression, or consequences of the incident.
    \item \textbf{Traffic Public Sentiment Analysis (T5)}: Given the feedback from the public regarding the proposed traffic proposal, the model should analyze, summarize, and discern the authentic demands of the public. This task facilitates a more comprehensive understanding for professionals of public sentiment, enabling targeted actions to be taken to fulfill the public's needs.
    \item \textbf{Traffic Safety Recommendation (T6)}: Given travel plans, such as weather conditions and road conditions, the model needs to provide reasonable safety traffic advice. Additionally, given an image, the model can point out the hidden security risks.
    \item \textbf{Traffic Sign Error Detection (T7)}: Given images containing traffic signs or lines on the road, the model needs to analyze whether the traffic signs are obstructed or defaced, whether traffic lines are designed reasonably, or if these lines need to be redrawn due to damage.
    \item \textbf{Traffic Road Occupation Detection (T8)}: Given images of roads, the model needs to analyze whether there is any illegal occupation of the road.
    
\end{itemize}

\paragraph{Transportation Knowledge Applying Tasks}
\begin{itemize}
    \item \textbf{Traffic Emergency Plan Generation (T9)}: Given an urgent description of a traffic accident or malfunction, the model should generate targeted emergency response plans.
    \item \textbf{Traffic Safety Education Copy Generation (T10)}: When provided with specific target audiences, the model should generate targeted educational materials.
\end{itemize}

\subsection{Data Collection}\label{sec:3-data}
In this section, a comprehensive exposition is presented regarding the process of data collection, including the data sources, data processing procedures, and ultimately, culminating in an overview of the acquired data.

\paragraph{Data Sources}
The aforementioned tasks primarily involve two modalities: text and images. 
For textual data, the primary source of our dataset is available on the internet. 
For instance, we have retrieved numerous examination papers related to the field of transportation for the source of question-and-answer data. 
The accident reports or public sentiment about specific regulations are predominantly sourced from news websites and municipal management platforms.
Additionally, specialized articles, such as emergency response plans, are primarily obtained from relevant sections of various institutional websites.
As for image data, we employ keyword-based queries to retrieve and select images from online repositories, ensuring conformity with predefined criteria. Simultaneously, the text from image titles or title links is preserved for further analysis.

\paragraph{Data Processing}
The formats of collected data are various, including Microsoft Word documents, PDFs, PNGs, JPGs, or Web pages.
We employ relevant Python toolkits (\textit{e.g.}, \texttt{pdfplumber}\footnote{\url{https://pypi.org/project/pdfplumber/}}, \texttt{pypdf2}\footnote{\url{https://pypi.org/project/PyPDF2/}}, \texttt{python-docx}\footnote{\url{https://pypi.org/project/python-docx/}}) to extract text and preprocess it into the appropriate format for the designated tasks. In cases where automated extraction is not feasible, we seek the relevant professionals to process it manually.
Additionally, we take measures to eliminate sensitive information from the data, including but not limited to personal phone numbers, ID numbers, emails, and detailed home addresses, safeguarding privacy. Furthermore, we ensure that each piece of data has undergone meticulous manual verification to guarantee alignment with the specified task, accuracy of answers, and coherence of sentences.

\begin{figure}[t]
    \centering
    \includegraphics[width=0.8\linewidth]{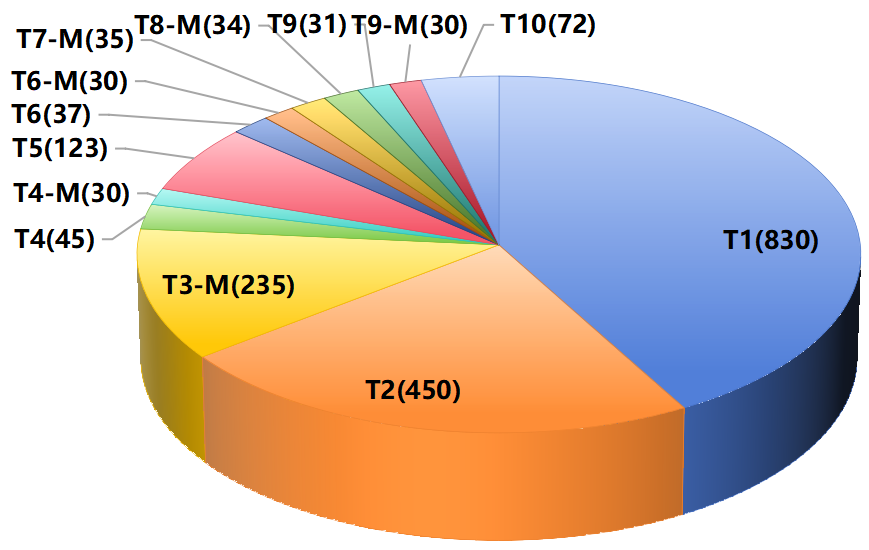}
    \caption{The distribution of data amounts for different tasks.
    ``-M'' means the multimodal dataset.
    }
    \label{fig:data}
\end{figure}

\paragraph{Data Overview}
Following data processing and manual verification, we obtain the final dataset corresponding to each task. 
Due to variations in task difficulty, the amount of data instances is different across tasks. A detailed data distribution is shown in Figure \ref{fig:data}.
Additionally, according to the involved modalities of different tasks (refer to the fourth column in Table \ref{table:tasks}), the entire dataset can be divided into two parts, the text-only dataset and the multimodal dataset, which will be utilized to evaluate LLMs and MLLMs respectively.
The input for the text-only dataset is a text question and the input for the multimodal dataset is an image with a question.
We have listed some examples for each task in the Appdenix \ref{sec:appendix-examples}.

\subsection{Evaluation}\label{sec:3-evaluation}
For the evaluation of each task, we first extract the answer from the model-generated prediction and then compute the corresponding metric scores according to the golden answer.

\paragraph{Answer Extraction}
For MLC and TF, some models generate answers that include content other than ``A/B/C/D'' or ``True/False''. It is imperative to extract the options from the generated answers in such cases. Moreover, we do not conduct extraction to other question types.

\paragraph{Different Metrics}
\begin{itemize}
    \item \textbf{Accuracy:} For MLC and TF, there are the gold answers for each query (T1$\sim$T3). Therefore, we calculate the accuracy of the extracted answer according to the gold answer. Additionally, we also calculate the format error rate of model-generated answers.
    \item \textbf{ROUGE:} For the questions of Generation type (T4$\sim$T10), we calculate the ROUGE-Chinese-L\footnote{\url{https://pypi.org/project/ROUGE-chinese/}} score between the predicted answer and the reference answer. ROUGE-L is a commonly used metric in generation tasks.
    \item \textbf{GPT-4-Eval:} Since the reference answers for some tasks (T4$\sim$T10) are not unique, we also utilize GPT-4\footnote{The \texttt{0613} version.} to evaluate the model-generated answers for accuracy, redundancy, fluency, and completeness. The example instruction that we designed is presented in Appendix \ref{sec:appendix-instruction-gpt-4}.
\end{itemize}

\begin{table*}[t]
    \centering
    \resizebox*{\linewidth}{!}{
    \begin{tabular}{l|ccccc}
    \bottomrule
       \textbf{Model}  & \textbf{Parameters} & \textbf{SFT} & \textbf{RLHF} & \textbf{Access} & \textbf{BaseModel} \\
    \hline
         \textit{\textbf{Large Language Models}} &  &  &  &  & \\
         ChatGLM3-6B (\citeauthor{zeng2023glm130b}) & 6B & \ding{51} & \ding{55} & Weights & ChatGLM \\
         Qwen-7B-Chat (\citeauthor{qwen}) & 7B &  \ding{51} & \ding{55} & Weights & Qwen-7B \\
         Qwen-14B-Chat (\citeauthor{qwen}) & 14B & \ding{51} & \ding{55} & Weights & Qwen-14B \\
         Baichuan2-13B-Chat (\citeauthor{baichuan2023baichuan2}) & 13B & \ding{51} & \ding{55} & Weights & Baichuan2-13B-Base \\ 
         InternLM-Chat-7B (\citeauthor{2023internlm}) & 7B &   \ding{51} & \ding{51}  & Weights & InternLM-7B \\ 
         InternLM-Chat-20B (\citeauthor{2023internlm}) & 20B &   \ding{51} & \ding{51}  & Weights & InternLM-20B \\
         Yi-6B-Chat & 6B & \ding{51} & \ding{55} & Weights & Yi-6B \\ 
         LLaMa2-Chinese-13B-Chat-ms & 13B & \ding{51} & \ding{55} & Weights & LLaMa2-13B \\ 
         GPT-4 & / & \ding{51} & \ding{51} & API & /\\ 
    \hline
         \textit{\textbf{Multimodal Large Language Models}}&  &  &  &  & \\
         VisualGLM (\citeauthor{zeng2023glm130b}) & 7.8B & \ding{51} & \ding{55} & Weights & ChatGLM-6B + BLIP2-Qformer \\
         mPLUG-Owl2 (\citeauthor{ye2023mplugowl2}) & 8.2B & \ding{51} & \ding{55} & Weights & LLaMa-7B + CLIP ViT-L/14\\
         Qwen-VL-Chat (\citeauthor{Qwen-VL}) & 9.6B & \ding{51} & \ding{55} & Weights & Qwen-7B + ViT-G/16\\
         Chinese-LLaVa-Cllama2 & 7.3B &  \ding{51} & \ding{55} & Weights  & LLaVa + Chinese-LLaMa2-7B\\
         Chinese-LLaVa-Baichuan & 7.3B &  \ding{51} & \ding{55} & Weights & LLaVa + Baichuan-7B\\
         InternLM-XComposer-7B  (\citeauthor{zhang2023internlmxcomposer}) & 8B &  \ding{51} & \ding{55} & Weights & InternLM-Chat-7B + EVA-CLIP\\
         LLaVa-v1.5-13B (\citeauthor{liu2023improved})  & 13.4B & \ding{51} & \ding{55} & Weights & Vicuna-v1.5-13B + CLIP ViT-L/14\\
    \hline
         \textit{\textbf{Transportation-domain Models}}&  &  &  &  & \\
         TransGPT (\citeauthor{TransGPT}) & 7B &  \ding{51} & \ding{55} & Weights  & LLaMa-7B\\
         TransGPT-MM (\citeauthor{TransGPTMM}) & 7.8B &  \ding{51} & \ding{55} & Weights  & VisualGLM\\
         TransCore-M  & 13.4B &  \ding{51} & \ding{55} & Weights  & PCITransGPT-13B + CLIP ViT/L-14\\
         
    \toprule
    \end{tabular}
    }
    \caption{Models tested on \benchmark. We classify these models by different modalities and we list the open-source models TransGPT(-MM) and TransCore-M in the transportation domain separately.}
    \label{table:model_list}
\end{table*}

\section{Experiments}

\subsection{Selected Models}
\label{sec:4-models}
We evaluate a substantial number of models on our \benchmark. According to modalities and domains, they are primarily categorized into three groups: Large Language Models (LLMs), Multimodal Large Language Models (MLLMs), and Transportation-domain Models (T-LMs).
Table \ref{table:model_list} presents an overall model list.
Specifically, for LLMs, we select some common models, such as ChatGLM3 \citep{zeng2023glm130b}, Qwen-7/14B-Chat \citep{qwen}, Baichuan2-13B-Chat \citep{baichuan2023baichuan2}, InternLM-Chat-7/20B \cite{2023internlm}, Yi-6B-Chat\footnote{\url{https://www.modelscope.cn/models/01ai/Yi-6B-Chat/summary}}, and 
LLaMa2-Chinese-13B-Chat-ms\footnote{\url{https://www.modelscope.cn/models/modelscope/Llama2-Chinese-13b-Chat-ms/summary}}.
We also evaluate GPT-4\footnote{\url{https://chat.openai.com/}} on our \benchmark.
For MLLMs, we pick out some models that support Chinese, such as VisualGLM \citep{zeng2023glm130b}, mPLUG-Owl2 \citeauthor{ye2023mplugowl2}, Qwen-VL-Chat \citep{Qwen-VL}, Chinese-LLaVa-Cllama2\footnote{\url{https://huggingface.co/LinkSoul/Chinese-LLaVA-Cllama2}}/Baichuan\footnote{\url{https://huggingface.co/LinkSoul/Chinese-LLaVA-Baichuan}}, InternLM-XComposer-7B  \citep{zhang2023internlmxcomposer}, and LLaVa-v1.5-13B \citep{liu2023improved}.
Moreover, we also evaluate TransGPT(-MM)~\cite{TransGPT, TransGPTMM} and TransCore-M\footnote{\url{https://huggingface.co/PCIResearch/TransCore-M}}, the open-sourced models in the transportation domain.
The more detailed information is shown in Table \ref{table:model_list}.

\subsection{Experimental Settings}
\label{sec:4-settings}
We set the input token length limit to 2048 and the output token length to 1024. Right truncation is performed for input prompts exceeding the length limitation.
For all open-sourced models, we set the officially recommended decoding strategy for each model.
Additionally, we evaluate all models in the zero-shot setting.
We utilize the text-only dataset and the multimodal dataset to evaluate LLMs and MLLMs respectively.

\begin{table*}[t]
    \centering
    \resizebox*{\linewidth}{!}{
    \begin{tabular}{c|cccccccc}
    \bottomrule
    \textbf{Models} & \textbf{T1} & \textbf{T2}  & \textbf{T4} & \textbf{T5} & \textbf{T6}  & \textbf{T9}  & \textbf{T10} & \textbf{SUM}\\
    \hline
    GPT-4 	&	\textbf{81.33}$_{\textbf{(0.00)}}$	&	80.89$_{\textbf{(0.00)}}$ 	&	21.2/\textbf{88.6}  &	44.3/\textbf{99.5} 	&	10.6/97.6 	&	19.4/\textbf{93.6}		&	\textbf{18.1}/\textbf{95.4} & \textbf{750.52}	\\
    Qwen-14B-Chat 	&	{80.12}$_{(0.36)}$	&	84.89$_{(0.22)}$ 	&	20.2/82.6 	&	39.2/97.5 	&	12.6/96.0 	&	20.8/87.7 		&	16.4/89.4 & 727.34	\\
    Yi-6B-Chat	&	79.16$_{(11.08)}$ 	&	\textbf{87.78}$_{(7.11)}$ 	&	14.8/85.6 	&	39.5/97.8 	&	7.5/\textbf{98.0} 	&	17.3/85.4 		&	11.4/92.7 	& 717.00\\
    Baichuan2-13B-Chat	&	69.04$_\textbf{\textbf{{(0.00)}}}$	&	77.11$_\textbf{{(0.00)}}$ 	&	\textbf{22.9}/83.8 	&	35.9/97.3 	&	9.0/97.3 	&	18.8/93.0		&	13.8/93.9 & 	711.72\\
    Qwen-7B-Chat	&	71.81$_{(4.94)}$ 	&	82.22$_{(2.67)}$ 	&	17.7/79.7 	&	39.5/97.2 	&	12.7/96.9 	&	19.9/83.4 		&	16.5/84.9 	& 702.44\\
    ChatGLM3-6B	&	63.98$_{(7.95)}$ 	&	71.56$_{(7.56)}$ 	&	21.0/83.5 	&	36.1/96.4 	&	9.1/95.1 	&	19.0/89.6 		&	14.9/89.1  & 689.43	\\
    InternLM-Chat-20B	&	62.89$_\textbf{{(0.00)}}$	&	76.44$_\textbf{{(0.00)}}$ 	&	11.0/50.8 	&	\textbf{49.6}/95.7 	&	12.1/96.9 	&	\textbf{22.2}/90.4 		&	17.2/92.0 	& 677.21\\
    InternLM-Chat-7B	&	62.65$_{(0.12)}$ 	&	66.00$_\textbf{{(0.00)}}$ 	&	18.7/72.7 	&	37.8/87.6 	&	\textbf{15.4}/88.0 	&	19.9/81.1 		&	17.5/89.6	& 656.81\\
    LLaMa2-Chinese-13B-Chat-ms	&	49.64$_{(2.05)}$ 	&	62.89$_{(3.33)}$ 	&	16.1/75.5 	&	35.6/94.0 	&	10.1/88.3	&	20.4/84.1 		&	14.1/77.1  & 627.65	\\

    \toprule
    \end{tabular}
    }
    \caption{
    The evaluation results of LLMs on the text-only dataset of our \benchmark.
    For \textbf{T1} and \textbf{T2} tasks, the values of Accuracy are listed and the format error rate is placed in the bottom right corner marker.
    ``xx/yy'' in the T4$\sim$T10 columns represents the values of the ``ROUGE/GPT-4-Eval'' metrics.
    The larger the value of all metrics except the format error rate, the better the performance.
    ``SUM'' is the sum of all values of different tasks, and we list all results according to the value of ``SUM'' from largest to smallest.
    Results highlighted in \textbf{bold} represent the best result in each column.
    }
    \label{table:LLMs-res}
\end{table*}

\begin{table*}[t]
    \centering
    \resizebox*{\linewidth}{!}{
    \begin{tabular}{c|ccccccc}
    \bottomrule
    \textbf{Models} & \textbf{T3} & \textbf{T4}  & \textbf{T6} & \textbf{T7} & \textbf{T8} & \textbf{T9} & \textbf{SUM}\\
    \hline
    
    Qwen-VL-Chat	&	\textbf{54.47}$_\textbf{{(0.00)}}$	&	9.3/75.1	&	\textbf{15.3}/\textbf{86.7}	&	7.4/\textbf{70.5}	&	20.6/\textbf{85.9}	&	14.4/64.5		& \textbf{504.15}	\\
    InternLM-XComposer-7B	&	48.94$_\textbf{{(0.00)}}$	&	8.9/77.9		&	16.1/86.4	&	\textbf{10.5}/56.4	&	\textbf{32.7}/67.7	&	\textbf{19.7}/77.6		& 	502.76\\
    TransCore-M	&	46.81$_\textbf{{(0.00)}}$	&	8.0/\textbf{79.3}	&	11.6/82.1	&	7.2/60.8	&	13.2/80.3	&	19.1/77.6	& 486.01	\\
    LLaVa-v1.5-13B	&	48.94$_{(1.28)}$	&	10.3/67.4		&	14.0/79.3	&	6.5/54.4	&	15.9/67.6	&	18.3/77.9		& 460.51	\\
    Chinese-LLaVa-Baichuan	&	20.43$_{(80.85)}$	&	6.9/73.5		&	9.9/84.6	&	4.2/60.5	&	10.3/73.4 &	14.0/\textbf{82.0}		& 439.80	\\
    VisualGLM-6B	&	26.38$_{(79.15)}$	&	10.1/73.0		&	11.6/77.6	&	7.4/64.0	&	8.8/75.2	&	14.6/65.6		& 	434.18\\
    mPLUG-Owl2	&	40.43$_{(0.43)}$	&	\textbf{11.6}/64.0		&	14.8/71.1	&	8.8/48.3	&	22.7/60.8	&	14.9/70.4		& 	427.66\\
    Chinese-LLaVa-Cllama2	&	8.09$_{(88.94)}$	&	7.6/65.5	&	10.3/83.5	&	4.5/54.1	&	9.3/74.7	&	12.2/79.5		& 	409.39\\
    
    \toprule
    \end{tabular}
    }
    \caption{
    The evaluation results of MLLMs on the multimodal dataset of our \benchmark.
    For \textbf{T3} task, the values of Accuracy are listed and other pattern introduction is the same as Table \ref{table:LLMs-res}.
    }
    \label{table:MLLMs-res}
\end{table*}

\subsection{Main Results}
\label{sec:4-results}
The evaluation results of the selected models on our {\benchmark} are shown in Table \ref{table:LLMs-res} and Table \ref{table:MLLMs-res}. Please note that there are still some models whose results are still being tested, so only partial results have been listed.
Next, we will introduce the performance of LLMs and MLLMs separately.

\paragraph{Large Language Models}
Table \ref{table:LLMs-res} presents the evaluation results of LLMs on the text-only dataset of our \benchmark.
The values of ``SUM'' in the last column show that GPT-4 obtains the best performance on most tasks and overall evaluation and Qwen-14B-Chat ranks second.
Yi-6B-Chat also achieves outstanding performance on many tasks, such as the \textbf{T2} and \textbf{T6} tasks, ranking third.
Overall, it is promising that some LLMs perform well in memorizing, understanding, and applying transportation knowledge, but there’s still room for improvement.

\paragraph{Multimodal Large Language Models} 
The evaluation results of MLLMs on the multimodal dataset shown in Table \ref{table:MLLMs-res} present that Qwen-VL-chat performs excellently on the majority of tasks and ranks first as a whole. 
InternLM-XComposer-7B ranks second and LLaVa-v1.5-13B ranks third.
However, even the top-performing model in the \textbf{T3} task, Qwen-VL-chat, achieves only 54.47\% accuracy, indicating the poor capability of MLLMs in the multimodal transportation domain.

\subsection{Analysis}
\label{sec:4-analysis}

\paragraph{Different models have different instruction-conforming capacities in T1/T2/T3 tasks.}
According to the format error rate of T1/T2/T3 tasks listed in Table \ref{table:LLMs-res} and Table \ref{table:MLLMs-res}, we observe that the format error rate of GPT-4 and the InternLM series models are all zero, demonstrating the excellent instruction-following ability.
We speculate that the reason may be that these models have been trained with RLHF.

\paragraph{There is still much room for improvement for some tasks.}
Due to the varying difficulty of different tasks, the performance of the models also varies. Overall, the model performs poorly on difficult tasks, especially in all tasks of multimodal scenarios as shown in Table \ref{table:MLLMs-res}. This provides a guiding direction for the model to further adapt to the transportation field.

\paragraph{The BaseModel is a key factor affecting model performance.} 
The selection of BaseModel is critical to the overall model performance, as the model learns large-scale knowledge during the pre-training phase.
We can observe from Table \ref{table:LLMs-res} and Table \ref{table:MLLMs-res} that the performance of some small-scale models can even outperform that of many large-scale models, such as Yi-6B-Chat surpassing InternLM-Chat-20B, Qwen-7B-Chat surpassing LLaMa2-Chinese-13B-Chat-ms, Qwen-VL-Chat surpassing LLava-v1.5-13B, and so on.
Additionally, due to the limited amount of Chinese corpus learned by LLaMa during the pre-training stage, the performance of the LLaMa series models is unsatisfactory such as LLaMa2-Chinese-13B-Chat-ms and Chinese-LLaVa-Cllama2.
These results further demonstrate the importance of the BaseModel, which almost determines the upper limit of model performance.

\paragraph{Scaling up the model size improves the performance with the similar BaseModel.}
The results in Table \ref{table:LLMs-res} showcase that Qwen-14B-Chat exceeds Qwen-7B-Chat and InternLM-Chat-20B exceeds InternLM-Chat-7B, which indicates that expanding the model scale will further improve the model performance when the BaseModel is the model of the same series.

\section{Conclusion}
In this work, we propose {\benchmark}, a carefully designed and thorough evaluation benchmark for assessing (M)LLMs in the transportation domain.
Referring to the first three levels in Bloom's Taxonomy, we test the performance of various (M)LLMs in memorizing, understanding and applying transportation knowledge.
The experimental results show that although some models perform well in some tasks, there is still much room for improvement overall.
Additionally, we analyze the key factors affecting model performance, which is helpful for how to further improve model performance.
We hope the release of {\benchmark} can serve as a foundation for future research, thereby accelerating the implementation and application of (M)LLMs in the field of transportation.

Furthermore, due to the need to connect to external databases for some scenarios in the transportation domain, such as real-time road condition queries and traffic flow prediction, our {\benchmark} does not include this type of task. In future work, we will further test the ability of (M)LLMs as an agent to call relevant interfaces to achieve specified tasks.

\section*{Limitations}
First, the biggest limitation is data leakage as our data is collected from the Internet. Although the original format of the data is complex and various, it is still difficult to ensure that existing (M)LLMs have not been directly trained on relevant data. We will explore more effective methods to prevent data leakage and strive for a more fair evaluation.

Second, the evaluation of long text generation tasks is very difficult, and we used ROUGE-L and GPT-4-Eval to evaluate the model-generated predictions together in our work. Due to the non-uniqueness of the answers, it is still difficult to ensure that the same effect as manual evaluation can be achieved.

Moreover, due to time constraints and the large amount of existing open-source models, we only test a small portion of common models in this work. We will test more models in the future.

\bibliography{anthology,custom}
\bibliographystyle{acl_natbib}

\appendix

\section{Examples of Tasks}
\label{sec:appendix-examples}

\begin{figure*}[ht]
    \centering
    \includegraphics[width=\linewidth]{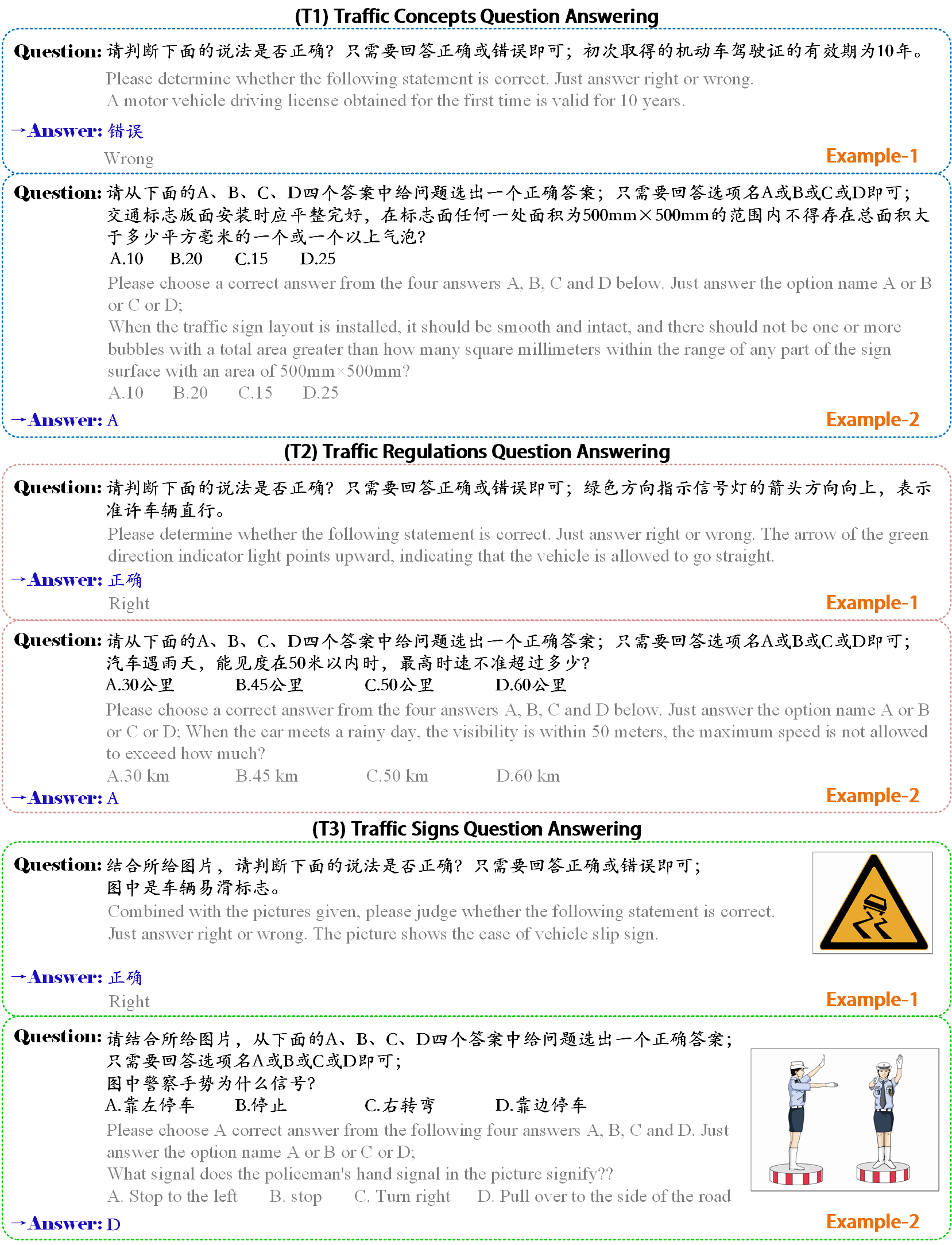}
    \caption{There are some examples for the \textbf{T1/T2/T3} tasks. The \textcolor{blue}{blue} text represents the answer, and the \textcolor{gray}{gray} text is the version of English. }
    \label{fig:a-b}
\end{figure*}

\begin{figure*}[ht]
    \centering
    \includegraphics[width=0.9\linewidth]{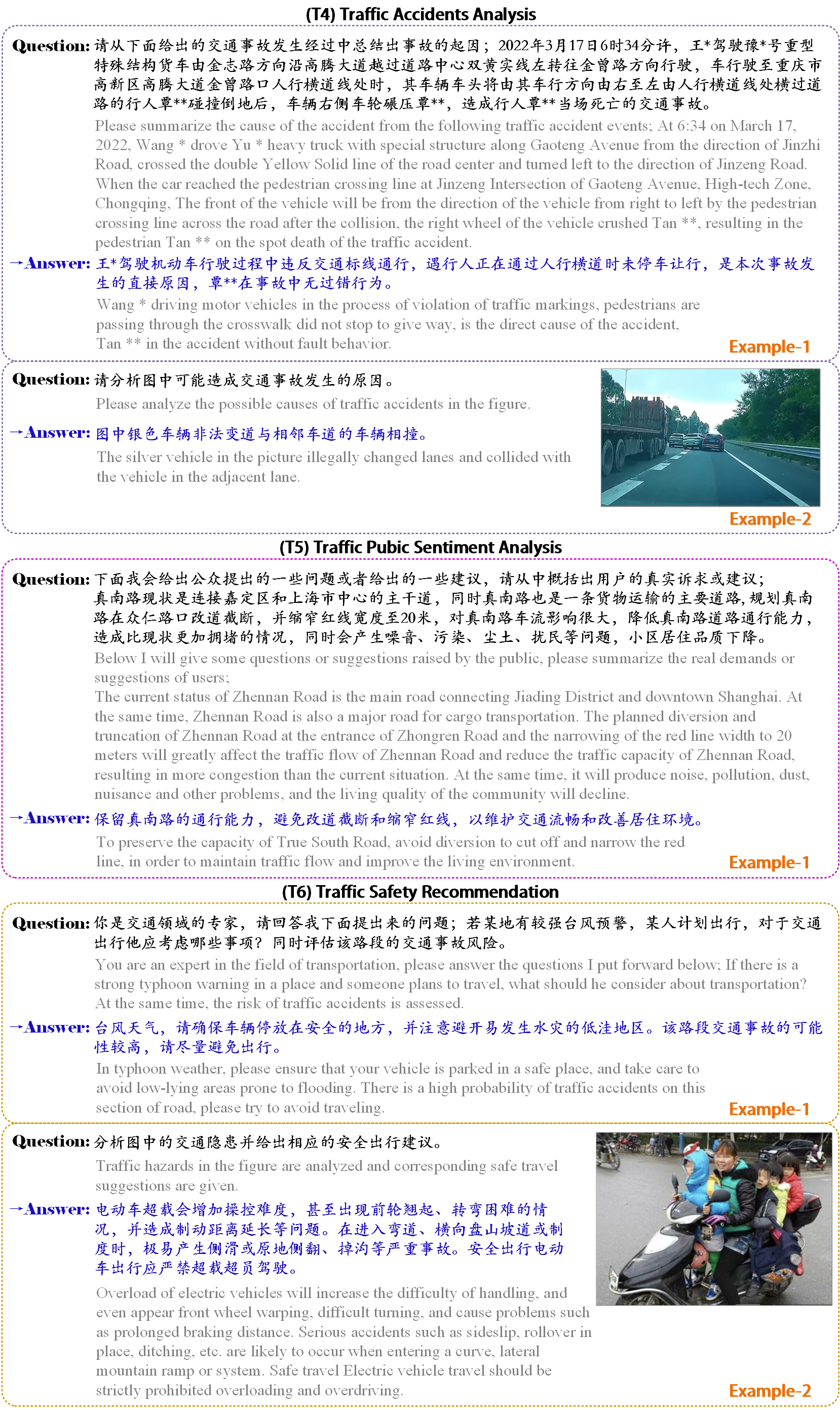}
    \caption{There are some examples for the \textbf{T4/T5/T6} tasks.}
    \label{fig:d-f}
\end{figure*}

\begin{figure*}[ht]
    \centering
    \includegraphics[width=0.89\linewidth]{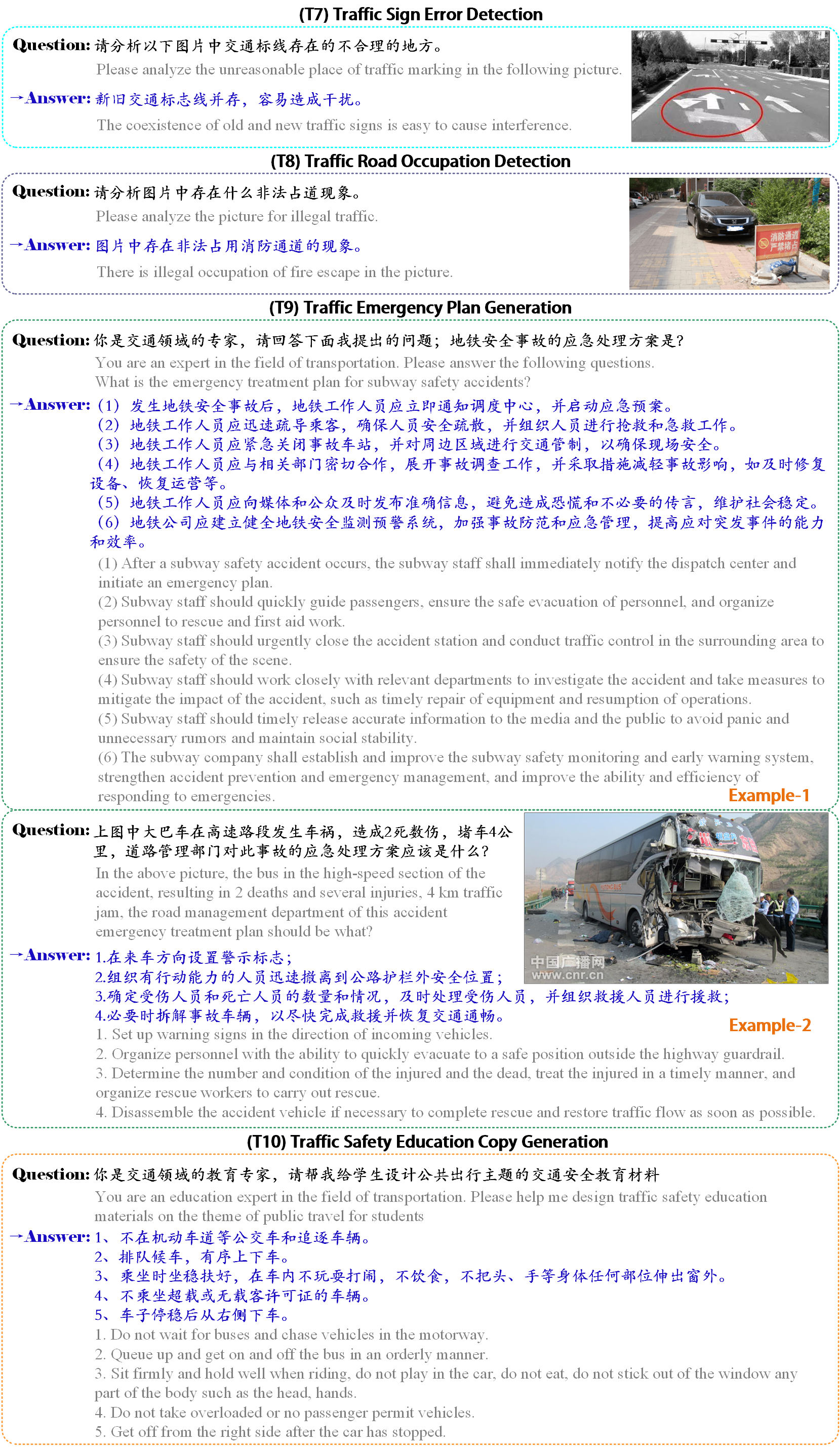}
    \caption{There are some examples for the \textbf{T7/T8/T9/T10} tasks.}
    \label{fig:g-k}
\end{figure*}

We list some examples for each task in Figure \ref{fig:a-b}, Figure \ref{fig:d-f}, and Figure \ref{fig:g-k}.

\section{An Example Instruction for GPT-4-Eval}
\label{sec:appendix-instruction-gpt-4}

We utilize GPT-4 to evaluate the model-generated answers for accuracy, redundancy, fluency, and completeness. The English version of the instruction is ``Below, I will give a question and a standard answer to the question, as well as an answer generated by the question-and-answer model. Since the answer is not unique, please judge the rationality of the answer generated by the question-and-answer model according to the reference answer given and combined with the actual situation, and it is necessary to consider the logic/accuracy/redundancy/fluency/integrity of the generated answer. The returned format is JSON, and the field is gpt4-score: The value is a decimal in the range of 0 to 1. Three decimal places are reserved after the decimal point. Question: xxx. Standard answer: xxx. The answer generated by the question-and-answer model: xxx.''

\end{document}